\documentclass[letterpaper]{article}
\usepackage{format/aaai-no-copy}
\usepackage{times}
\usepackage{helvet}
\usepackage{courier}
\usepackage{graphicx} 
\usepackage{subfigure} 

\usepackage{hyperref}
\usepackage{url}
\usepackage{color}
\usepackage{preamble}
\definecolor{mydarkblue}{rgb}{0,0.08,0.45}
\hypersetup{ %
    pdftitle={},
    pdfauthor={},
    pdfsubject={},
    pdfkeywords={},
    pdfborder=0 0 0,
    pdfpagemode=UseNone,
    colorlinks=true,
    linkcolor=mydarkblue,
    citecolor=mydarkblue,
    filecolor=mydarkblue,
    urlcolor=mydarkblue,
    pdfview=FitH}

\usepackage{amsmath, amsfonts, bm, lipsum, capt-of}
\usepackage{natbib, xcolor, wrapfig, booktabs, multirow, caption}
\usepackage{float}
\usepackage{datetime}
\usepackage[toc,page]{appendix}

\usepackage{tikz}
\usetikzlibrary{shapes.geometric,arrows,chains,matrix,positioning,scopes,calc}
\tikzstyle{mybox} = [draw=white, rectangle]

\def\ie{i.e.\ }
\def\eg{e.g.\ }

\let\emptyset 0

\newcommand{\procedurename}{ABCD}

\definecolor{WowColor}{rgb}{.75,0,.75}
\definecolor{SubtleColor}{rgb}{0,0,.50}

\newcounter{margincounter}

\begin{document}
%
\title{Automatic Construction and Natural-Language Description \\ of Nonparametric Regression Models}

\author{James Robert Lloyd\\
Department of Engineering\\
University of Cambridge\\
\And
David Duvenaud\\
Department of Engineering\\
University of Cambridge\\
\And
Roger Grosse\\
Brain and Cognitive Sciences\\
Massachusetts Institute of Technology\\
\AND
Joshua B. Tenenbaum\\
Brain and Cognitive Sciences\\
Massachusetts Institute of Technology\\
\And
Zoubin Ghahramani\\
Department of Engineering\\
University of Cambridge\\
}
\maketitle

\begin{abstract} 
\begin{quote}
This paper presents the beginnings of an automatic statistician, focusing on regression problems.
Our system explores an open-ended space of statistical models to discover a good explanation of a data set, and then produces a detailed report with figures and natural-language text.

Our approach treats unknown regression functions nonparametrically using Gaussian processes, which has two important consequences.
First, Gaussian processes can model functions in terms of high-level properties (e.g.\ smoothness, trends, periodicity, changepoints).
Taken together with the compositional structure of our language of models this allows us to automatically describe functions in simple terms.
Second, the use of flexible nonparametric models and a rich language for composing them in an open-ended manner also results in state-of-the-art extrapolation performance evaluated over 13 real time series data sets from various domains.
\end{quote}
\end{abstract} 

\allowdisplaybreaks

\section{Introduction}

Automating the process of statistical modeling would have a tremendous impact on fields that currently rely on expert statisticians, machine learning researchers, and data scientists.
While fitting simple models (such as linear regression) is largely automated by standard software packages, there has been little work on the automatic construction of flexible but interpretable models. 
What are the ingredients required for an artificial intelligence system to be able to perform statistical modeling automatically? 
In this paper we conjecture that the following ingredients may be useful for building an AI system for statistics, and we develop a working system which incorporates them:
\begin{itemize}
\item {\bf An open-ended language of models} expressive enough to capture many of the modeling assumptions and model composition techniques  applied by human statisticians to capture real-world phenomena
\item {\bf A search procedure} to efficiently explore the space of models spanned by the language
\item {\bf A principled method for evaluating models} in terms of their complexity and their degree of fit to the data
\item {\bf A procedure for automatically generating reports} which explain and visualize different factors underlying the data, make the chosen modeling assumptions explicit, and quantify how each component improves the predictive power of the model 
\end{itemize}

\begin{figure}[t]
\centering
\fbox{\includegraphics[trim=0cm 4.75cm 0cm 1.0cm, clip, width=0.98\columnwidth]{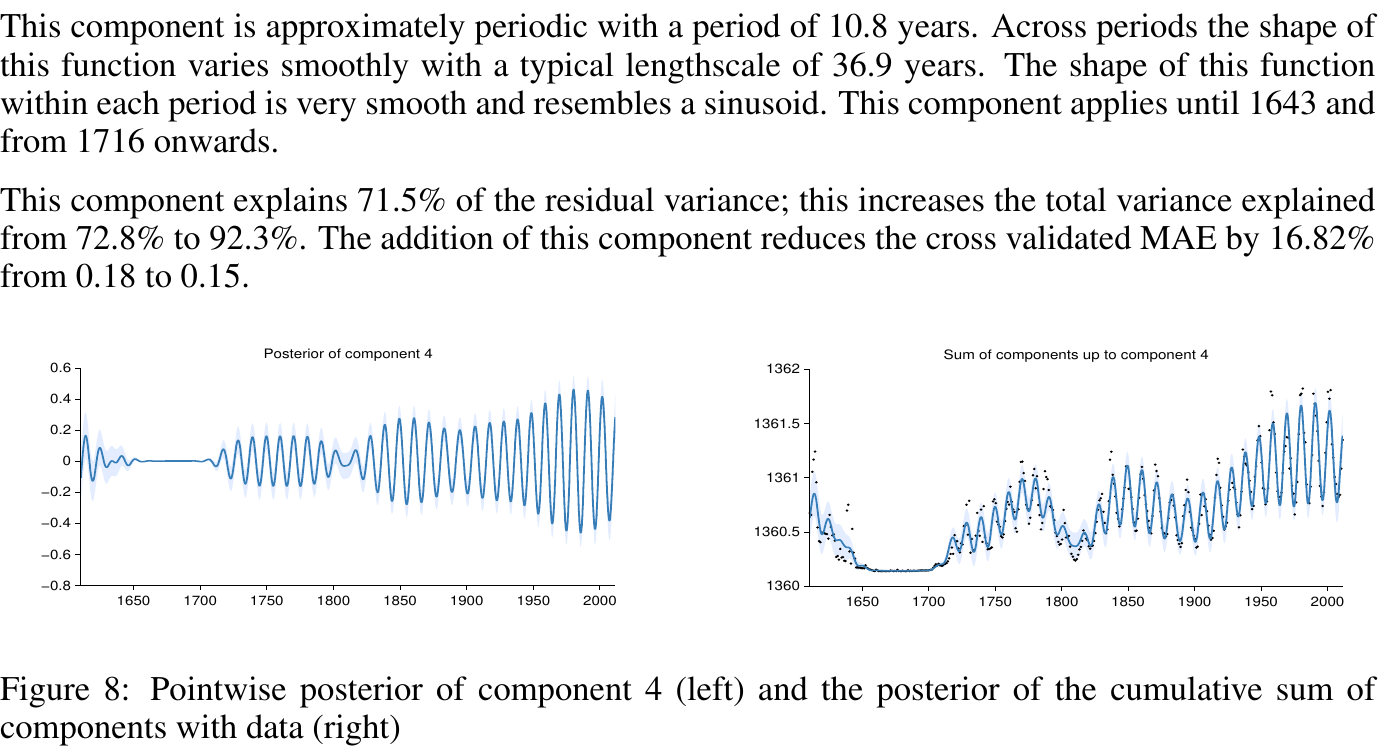}}
\caption{
Extract from an automatically-generated report describing the model components discovered by \procedurename{}.
This part of the report isolates and describes the approximately 11-year sunspot cycle, also noting its disappearance during the 16th century, a time known as the Maunder minimum \citep{lean1995reconstruction}.
}
\label{fig:periodic}
\end{figure}

In this paper we introduce a system for modeling time-series data containing the above ingredients which we call the Automatic Bayesian Covariance Discovery (ABCD) system.
The system defines an open-ended language of Gaussian process models via a compositional grammar.
The space is searched greedily, using marginal likelihood and the Bayesian Information Criterion (BIC) to evaluate models.
The  compositional structure of the language allows us to develop a method for automatically translating components of the model into natural-language descriptions of patterns in the data.

We show examples of automatically generated reports which highlight interpretable features discovered in a variety of data sets (\eg figure~\ref{fig:periodic}).
The supplementary material to this paper includes 13 complete reports automatically generated by ABCD.

Good statistical modeling requires not only interpretability but also predictive accuracy.
We compare ABCD against existing model construction techniques in terms of predictive performance at extrapolation, and we find state-of-the-art performance on 13 time series.

\section{A language of regression models}
\label{sec:improvements}

Regression consists of learning a function $f$ mapping from some input space $\mathcal{X}$ to some output space $\mathcal{Y}$.
We desire an expressive language which can represent both simple parametric forms of $f$ such as linear or polynomial and also complex nonparametric functions specified in terms of properties such as smoothness or periodicity.
Gaussian processes (\gp{}s) provide a very general and analytically tractable way of capturing both simple and complex functions. 

\gp{}s are distributions over functions such that any
finite set of function evaluations, $(f(x_1), f(x_2), \ldots
f(x_N))$, have a jointly Gaussian distribution
\citep{rasmussen38gaussian}. A \gp{} is completely specified by its
mean function, $\mu(x)=\mathbb{E}(f(x))$ and kernel (or covariance) function
$\kernel(x,x') = \Cov(f(x),f(x'))$.
It is common practice to assume zero mean,
since marginalizing over an unknown mean function can be equivalently
expressed as a zero-mean \gp{} with a new kernel. The structure of the
kernel captures high-level properties of the unknown function, $f$,
which in turn determines how the model generalizes or extrapolates to
new data.  We can therefore define a language of regression models by
specifying a language of kernels.

The elements of this language are a set of  base
kernels capturing different function properties, and a set of
composition rules which combine kernels to yield other valid kernels.
Our base kernels are white noise ($\kWN$), constant ($\kC$), linear ($\kLin$), squared exponential ($\kSE$) and periodic ($\kPer$), which on their own encode for uncorrelated noise, constant functions, linear functions, smooth functions and periodic functions respectively\footnote{Definitions of kernels are in the supplementary material.}.
The composition rules are addition and multiplication:
\begin{align}
(k_1 + k_2)(x,x') =& \,\, k_1(x,x') + k_2(x,x')\\
(k_1 \times k_2)(x,x') =& \,\, k_1(x,x') \times k_2(x,x')
\end{align}

Combining kernels using these operations can yield kernels encoding for richer structures such as approximate periodicity ($\kSE \times \kPer$) or smooth functions with linear trends ($\kSE + \kLin$).

This kernel composition framework (with different base kernels) was described by \citet{DuvLloGroetal13}.
We extend and adapt this framework in several ways.
In particular, we have found that incorporating changepoints into the language is essential for
realistic models of time series (\eg figure~\ref{fig:periodic}). 
We define changepoints through addition and multiplication with sigmoidal functions:
\begin{align}
\kCP(\kernel_1, \kernel_2) = \kernel_1 \times \boldsymbol\sigma + \kernel_2 \times \boldsymbol{\bar\sigma}
\label{eq:cp}
\end{align}
where $\boldsymbol\sigma = \sigma(x)\sigma(x')$ and $\boldsymbol{\bar\sigma} = (1-\sigma(x))(1-\sigma(x'))$.
We define changewindows $\kCW(\cdot,\cdot)$ similarly by replacing $\sigma(x)$ with a product of two sigmoids.

We also expanded and reparametrised the set of base kernels so that they were more amenable to automatic description (see section~\ref{sec:design} for details) and to extend the number of common regression models included in the language.
Table~\ref{table:motifs} lists common regression models that can be expressed by our language.
\begin{table}[ht]
\centering
\begin{tabular}{l|l}
Regression model & Kernel \\
\midrule
\gp{} smoothing & $\kSE + \kWN$ \\
Linear regression & $\kC + \kLin + \kWN$ \\
Multiple kernel learning & $\sum \kSE$ + \kWN\\
Trend, cyclical, irregular & $\sum \kSE + \sum \kPer$ + \kWN\\
Fourier decomposition* & $\kC + \sum \cos$ + \kWN\\
Sparse spectrum \gp{}s* & $\sum \cos$ + \kWN\\
Spectral mixture* & $\sum \SE \times \cos$ + \kWN\\
Changepoints* & \eg $\kCP(\kSE, \kSE) + \kWN$ \\
Heteroscedasticity* & \eg $\kSE + \kLin \times \kWN$
\end{tabular}
\caption{
Common regression models expressible in our language.
$\cos$ is a special case of our reparametrised $\kPer$.
* indicates a model that could not be expressed by the language used in \citet{DuvLloGroetal13}.
}
\label{table:motifs}
\end{table}

\section{Model Search and Evaluation}

As in \citet{DuvLloGroetal13} we explore the space of regression models using a greedy search.
We use the same search operators, but also include additional operators to incorporate changepoints; a complete list is contained in the supplementary material. 

After each model is proposed its kernel parameters are optimised by conjugate gradient descent.
We evaluate each optimized model, $M$, using the Bayesian Information Criterion (BIC) \citep{schwarz1978estimating}:
\begin{equation}
\textrm{BIC}(M) = -2 \log p(D\given M) + |M| \log n
\end{equation}
where $|M|$ is the number of kernel parameters, $p(D|M)$ is the marginal likelihood of the data, $D$, and $n$ is the number of data points.
BIC trades off model fit and complexity and implements what is known as ``Bayesian Occam's Razor'' \citep[e.g.][]{rasmussen2001occam,mackay2003information}.

\section{Automatic description of regression models}
\label{sec:description}

\paragraph{Overview}

In this section, we describe how \procedurename{} generates natural-language descriptions of the models found by the search procedure.
There are two main features of our language of \gp{} models that allow description to be performed automatically.

First, the sometimes complicated kernel expressions can be simplified into a sum of products.
A sum of kernels corresponds to a sum of functions so each product can be described separately.
Second, each kernel in a product modifies the resulting model in a consistent way.
Therefore, we can choose one kernel to be described as a noun, with all others described using adjectives or modifiers.

\paragraph{Sum of products normal form} 

We convert each kernel expression into a standard, simplified form.
We do this by first distributing all products of sums into a sum of products.
Next, we apply several simplifications to the kernel expression:
The product of two $\kSE$ kernels is another $\kSE$ with different parameters. Multiplying $\kWN$ by any stationary kernel ($\kC$, $\kWN$, $\kSE$, or $\kPer$) gives another $\kWN$ kernel. Multiplying any kernel by $\kC$ only changes the parameters of the original kernel.

After applying these rules, the kernel can as be written as a sum of terms of the form:
\begin{align*}
K \prod_m \kLin^{(m)} \prod_n \boldsymbol\sigma^{(n)},
\label{eq:sop}
\end{align*}
where $K$, if present, is one of \kWN, \kC, \kSE, $\prod_k \kPer^{(k)}$ or $\kSE \prod_k \kPer^{(k)}$
and $\prod_i\kernel^{(i)}$ denotes a product of kernels, each with different parameters.

\paragraph{Sums of kernels are sums of functions}
Formally, if $f_1(x) \dist \gp{}(0, \kernel_1)$ and independently $f_2(x) \dist \gp{}(0, \kernel_2)$ then ${f_1(x) + f_2(x) \dist \gp{}(0, \kernel_1 + \kernel_2)}$.
This lets us describe each product of kernels separately.

\paragraph{Each kernel in a product modifies a model in a consistent way}
This allows us to describe the contribution of each kernel as a modifier of a noun phrase.
These descriptions are summarised in table~\ref{table:modifiers} and justified below:

\begin{itemize}
\item {\bf Multiplication by $\kSE$} removes long range correlations from a model since $\kSE(x,x')$ decreases monotonically to 0 as $|x - x'|$ increases.
This will convert any global correlation structure into local correlation only.
\item {\bf Multiplication by $\kLin$} is equivalent to multiplying the function being modeled by a linear function.
If $f(x) \dist \gp{}(0, \kernel)$, then $xf(x) \dist \gp{}\left(0, k \times \kLin \right)$.
This causes the standard deviation of the model to vary linearly without affecting the correlation.
\item {\bf Multiplication by $\boldsymbol\sigma$} is equivalent to multiplying the function being modeled by a sigmoid which means that the function goes to zero before or after some point.
\item {\bf Multiplication by $\kPer$}
modifies the correlation structure in the same way as multiplying the function by an independent periodic function.
Formally, if ${f_1(x) \dist \gp{}(0, \kernel_1)}$ and ${f_2(x) \dist \gp{}(0, \kernel_2)}$ then
\begin{align}
{\textrm{Cov} \left[f_1(x)f_2(x), f_1(x')f_2(x') \right] = k_1(x,x')k_2(x,x')}.\nonumber
\end{align}
\end{itemize}

\begin{table}[ht]
\centering
\begin{tabular}{l|l}
Kernel & Postmodifier phrase \\
\midrule
$\kSE$  & whose shape changes smoothly \\
$\kPer$ & modulated by a periodic function \\
$\kLin$ & with linearly varying amplitude \\
$\prod_k \kLin^{(k)}$ & with polynomially varying amplitude \\
$\prod_k \boldsymbol{\sigma}^{(k)}$ & which applies until / from [changepoint] \\
\end{tabular}
\caption{
Postmodifier descriptions of each kernel
}
\label{table:modifiers}
\end{table}

\paragraph{Constructing a complete description of a product of kernels}
We choose one kernel to act as a noun which is then described by the functions it encodes for when unmodified (see table~\ref{table:nouns}).
Modifiers corresponding to the other kernels in the product are then appended to this description, forming a noun phrase of the form:
\begin{align*}
\textnormal{Determiner}	+	\textnormal{Premodifiers} +	\textnormal{Noun}	+	\textnormal{Postmodifiers}
\end{align*}

As an example, a kernel of the form $\kPer \times  \kLin \times \boldsymbol{\sigma}$ could be described as a
\begin{align*}
\underbrace{\kPer}_{\textnormal{\scriptsize periodic function}} \times 
\underbrace{\kLin}_{\textnormal{\scriptsize with linearly varying amplitude}} \times 
\underbrace{\boldsymbol{\sigma}}_{\textnormal{\scriptsize which applies until 1700.}}
\end{align*}
where $\kPer$ has been selected as the head noun.

\begin{table}[ht]
\centering
\begin{tabular}{l|l}
Kernel & Noun phrase \\
\midrule
$\kWN$  & uncorrelated noise \\
$\kC$   & constant \\
$\kSE$  & smooth function \\
$\kPer$ & periodic function \\
$\kLin$ & linear function \\
$\prod_k \kLin^{(k)}$ & polynomial \\
\end{tabular}
\caption{
Noun phrase descriptions of each kernel
}
\label{table:nouns}
\end{table}

\paragraph{Refinements to the descriptions}

There are a number of ways in which the descriptions of the kernels can be made more interpretable and informative:
\begin{itemize}
  \item Which kernel is chosen as the head noun can change the interpretability of a description.
  \item Descriptions can change qualitatively according to kernel parameters \eg `a rapidly varying smooth function'.
  \item Descriptions can include kernel parameters \eg `modulated by a periodic function with a period of [period]'.
  \item Descriptions can include extra information calculated from data \eg `a linearly increasing function'.
  \item Some kernels can be described as premodifiers \eg `an approximately periodic function'.
\end{itemize}

The reports in the supplementary material and in section~\ref{sec:examples} include some of these refinements.
For example, the head noun is chosen according to the following ordering:
\begin{align*}
\kPer > \kWN, \kSE, \kC > \prod_m \kLin^{(m)} > \prod_n \boldsymbol\sigma^{(n)}
\end{align*}
\ie $\kPer$ is always chosen as the head noun when present.
The parameters and design choices of these refinements have been chosen by our best judgement, but learning these parameters objectively from expert statisticians would be an interesting area for future study.

\paragraph{Ordering additive components}

The reports generated by \procedurename{} attempt to present the most interesting or important features of a data set first.
As a heuristic, we order components by always adding next the component which most reduces the 10-fold cross-validated mean absolute error.

\subsection{Worked example}

Suppose we start with a kernel of the form
\begin{align*}
\kSE \times (\kWN \times \kLin + \kCP(\kC, \kPer)).
\end{align*}
This is converted to a sum of products:
\begin{align*}
\kSE \times \kWN \times \kLin + \kSE \times \kC \times \boldsymbol{\sigma} + \kSE \times \kPer \times \boldsymbol{\bar\sigma}.
\end{align*}
which is simplified to
\begin{align*}
\kWN \times \kLin + \kSE \times \boldsymbol{\sigma} + \kSE \times \kPer \times \boldsymbol{\bar\sigma}.
\end{align*}

To describe the first component, the head noun description for $\kWN$, `uncorrelated noise', is concatenated with a modifier for $\kLin$, `with linearly increasing amplitude'.
The second component is described as `A smooth function with a lengthscale of [lengthscale] [units]', corresponding to the $\kSE$, `which applies until [changepoint]', which corresponds to the $\boldsymbol{\sigma}$.
Finally, the third component is described as `An approximately periodic function with a period of [period] [units] which applies from [changepoint]'.

\section{Example descriptions of time series}
\label{sec:examples}
We demonstrate the ability of our procedure to discover and describe a variety of patterns on two time series.
Full automatically-generated reports for 13 data sets are provided as supplementary material.

\subsection{Summarizing 400 Years of Solar Activity}
\label{sec:solar}

\begin{figure}[h]
\centering
\includegraphics[trim=0.2cm 18.0cm 8cm 2cm, clip, width=0.98\columnwidth]{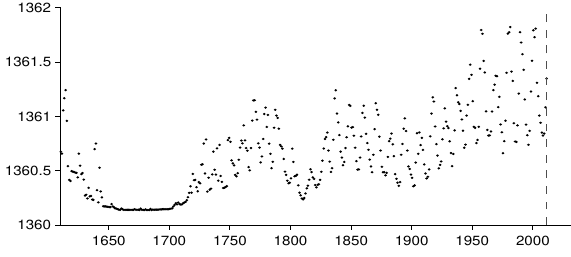}
\caption{
Solar irradiance data.}
\label{fig:solar}
\end{figure}

We show excerpts from the report automatically generated on annual solar irradiation data from 1610 to 2011 (figure~\ref{fig:solar}).
This time series has two pertinent features: a roughly 11-year cycle of solar activity, and a period lasting from 1645 to 1715 with much smaller variance than the rest of the dataset.
This flat region corresponds to the Maunder minimum, a period in which sunspots were extremely rare \citep{lean1995reconstruction}.
\procedurename{} clearly identifies these two features, as discussed below.

\begin{figure}[h]
\centering
\fbox{\includegraphics[trim=0cm 10.8cm 0cm 6.3cm, clip, width=0.98\columnwidth]{solarpages/pg_0002-crop}}
\caption{
Automatically generated descriptions of the components discovered by \procedurename{} on the solar irradiance data set.
The dataset has been decomposed into diverse structures with simple descriptions.}
\label{fig:exec}
\end{figure}
\begin{figure}[ht]
\centering
\fbox{\includegraphics[trim=0cm 4.75cm 0cm 0.7cm, clip, width=0.98\columnwidth]{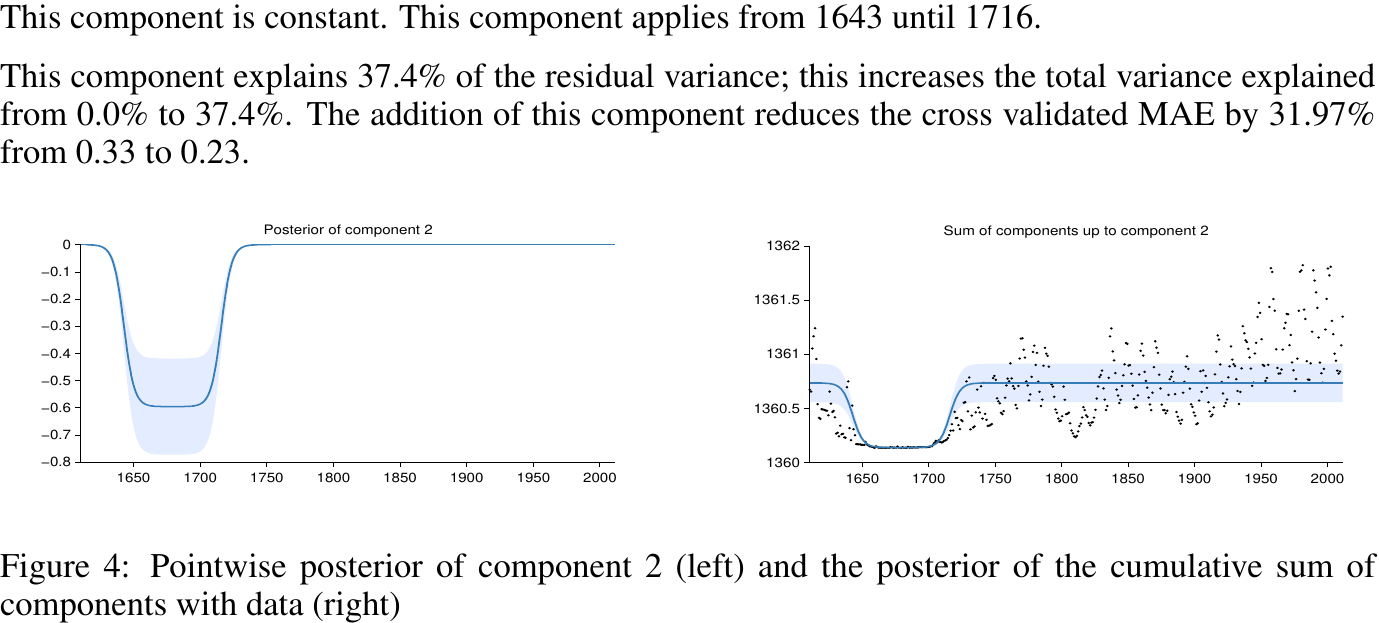}}
\caption{One of the learned components corresponds to the Maunder minimum.}
\label{fig:maunder}
\end{figure}
\begin{figure}[h!]
\centering
\fbox{\includegraphics[trim=0cm 4.75cm 0cm 1cm, clip, width=0.98\columnwidth]{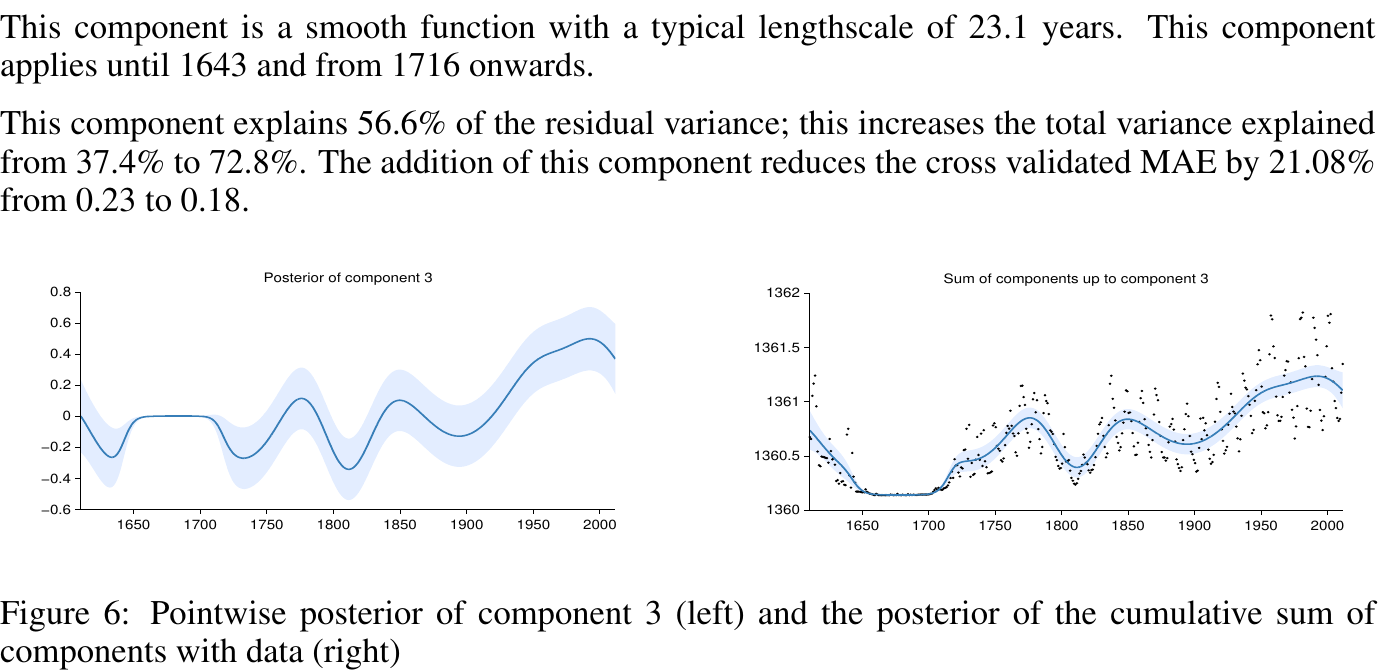}}
\caption{Characterizing the medium-term smoothness of solar activity levels.  By allowing other components to explain the periodicity, noise, and the Maunder minimum, \procedurename{} can isolate the part of the signal best explained by a slowly-varying trend.}
\label{fig:smooth}
\end{figure}

Figure \ref{fig:exec} shows the natural-language summaries of the top four components chosen by \procedurename{}.
From these short summaries, we can see that our system has identified the Maunder minimum (second component) and 11-year solar cycle (fourth component).
These components are visualized in figures~\ref{fig:maunder} and \ref{fig:periodic}, respectively. 
The third component corresponds to long-term trends, as visualized in figure~\ref{fig:smooth}.

\subsection{Finding heteroscedasticity in air traffic data}
\label{sec:airline}

\begin{figure}[h]
\centering
\includegraphics[trim=0.4cm 16.8cm 8cm 2cm, clip, width=0.98\columnwidth]{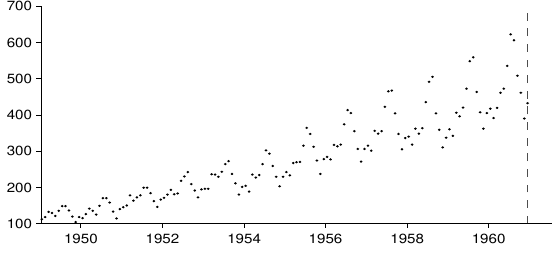}
\caption{
International airline passenger monthly volume \citep[e.g.][]{box2013time}.}
\label{fig:airline}
\end{figure}

\begin{figure}[h]
\centering
\fbox{\includegraphics[trim=0cm 6.8cm 0cm 6cm, clip, width=0.98\columnwidth]{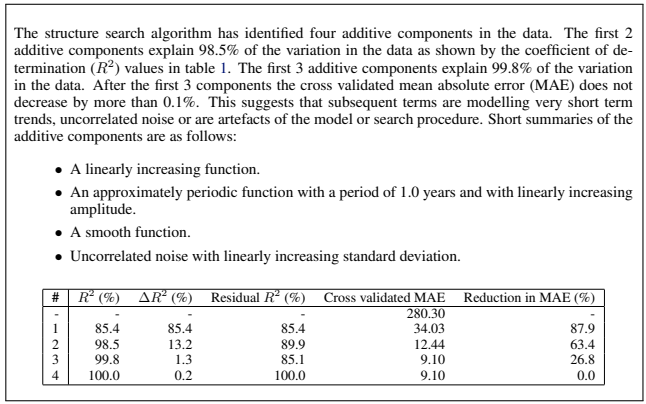}}
\caption{
Short descriptions and summary statistics for the four components of the airline model.}
\label{fig:exec-airline}
\end{figure}

Next, we present the analysis generated by our procedure on international airline passenger data (figure~\ref{fig:airline}).
The model constructed by \procedurename{} has four components: $\kLin + \kSE \times \kPer \times \kLin + \kSE + \kWN \times \kLin$, with descriptions given in figure~\ref{fig:exec-airline}.

\begin{figure}[h]
\centering
\fbox{\includegraphics[trim=0cm 4.75cm 0cm 0cm, clip, width=0.98\columnwidth]{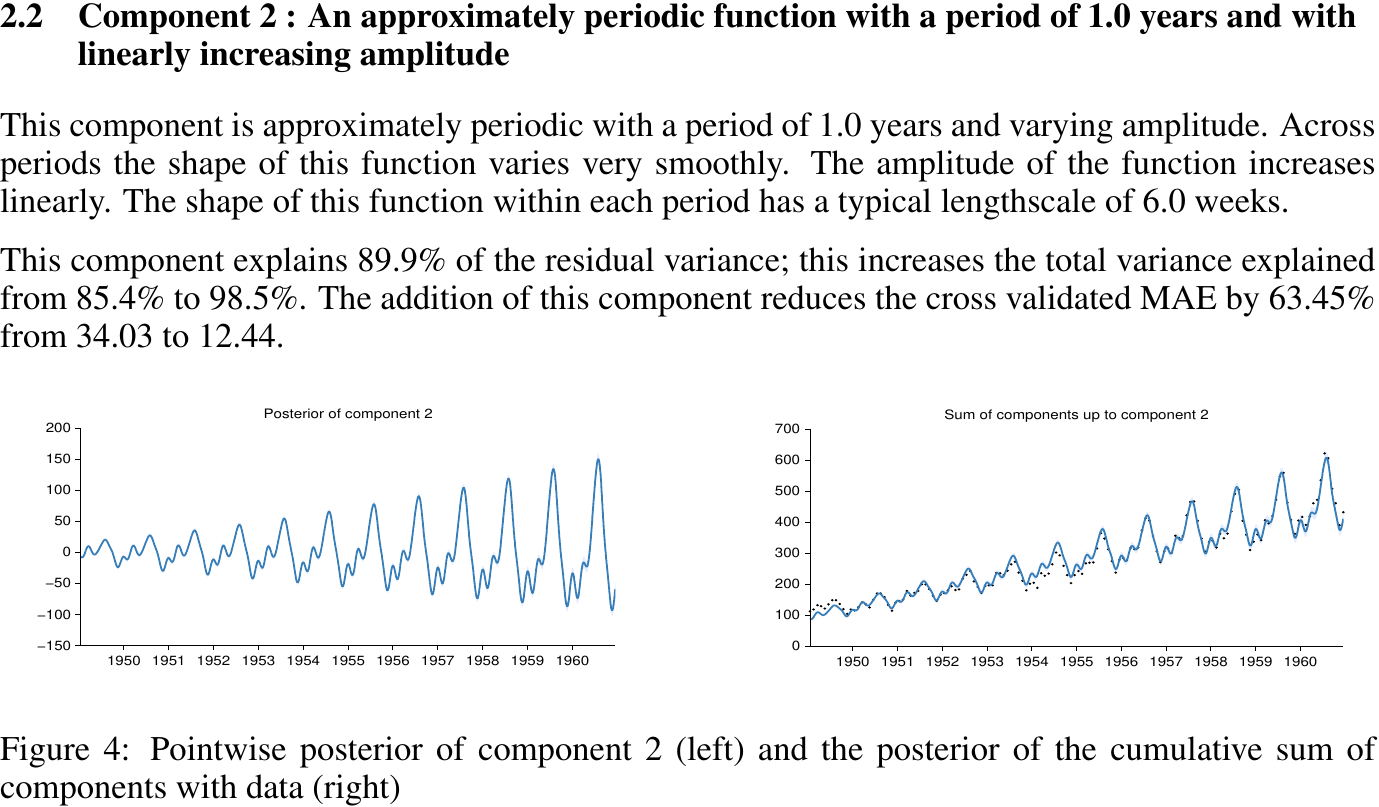}}
\caption{Capturing non-stationary periodicity in the airline data}
\label{fig:lin_periodic}
\end{figure}

\begin{figure}[h]
\centering
\fbox{\includegraphics[trim=0cm 0cm 0cm 0cm, clip, width=0.98\columnwidth]{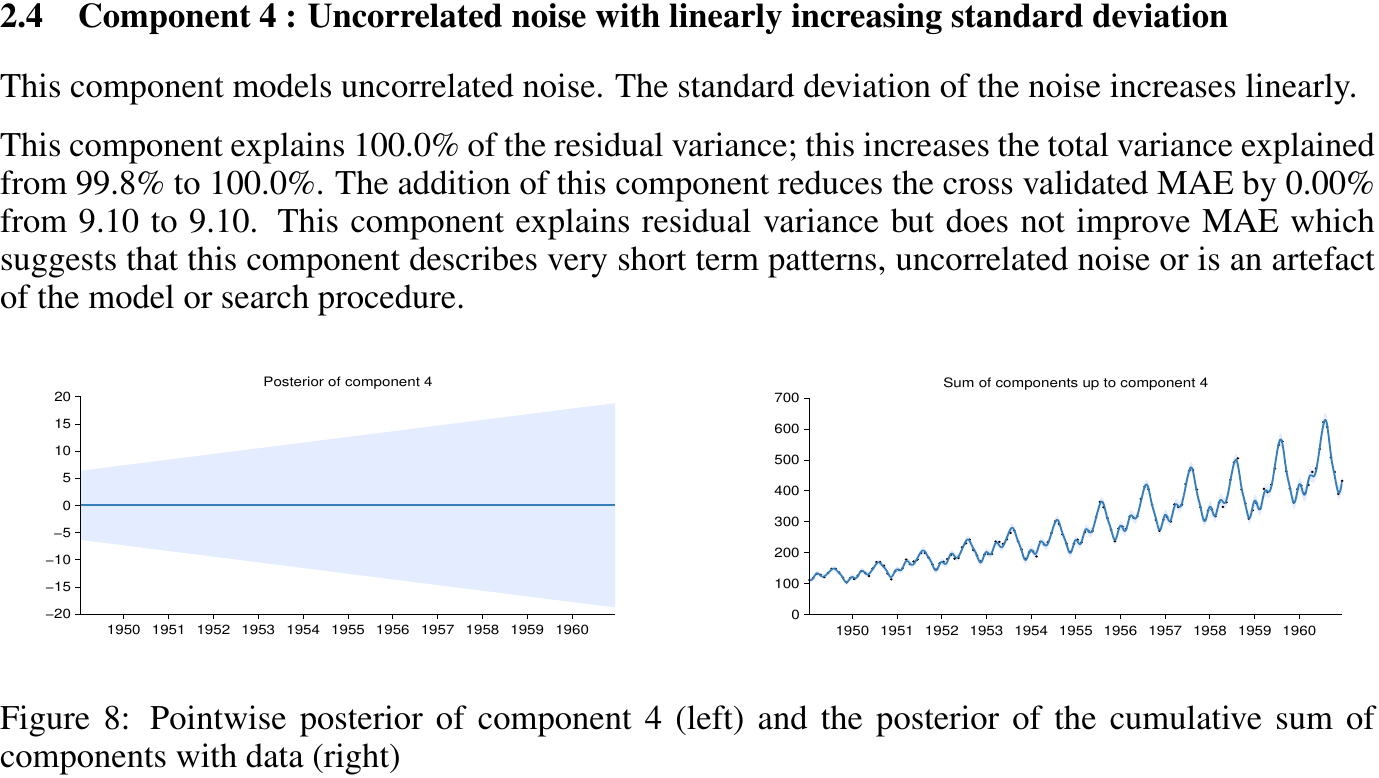}}
\caption{Modeling heteroscedasticity}
\label{fig:heteroscedastic}
\end{figure}

The second component (figure~\ref{fig:lin_periodic}) is accurately described as approximately ($\kSE$) periodic ($\kPer$) with linearly increasing amplitude ($\kLin$).
By multiplying a white noise kernel by a linear kernel, the model is able to express heteroscedasticity (figure~\ref{fig:heteroscedastic}).

\subsection{Comparison to equation learning}
\label{sec:eqn-learning-comp}

We now compare the descriptions generated by \procedurename{} to parametric functions produced by an equation learning system.
We show equations produced by Eureqa \citep{Eureqa} for the data sets shown above, using the default mean absolute error performance metric.

The learned function for the solar irradiance data is
\begin{align*}
\textrm{Irradiance($t$)} = 1361 + \alpha\sin(\beta + \gamma t)\sin(\delta + \epsilon t^2 - \zeta t)
\end{align*}
where $t$ is time and constants are replaced with symbols for brevity.
This equation captures the constant offset of the data, and models the long-term trend with a product of sinusoids, but fails to capture the solar cycle or the Maunder minimum.

The learned function for the airline passenger data is
\begin{align*}
\textrm{Passengers($t$)} = \alpha t + \beta\cos(\gamma - \delta t)\textrm{logistic}(\epsilon t - \zeta) - \eta
\end{align*}
which captures the approximately linear trend, and the periodic component with approximately linearly (logistic) increasing amplitude.
However, the annual cycle is heavily approximated by a sinusoid and the model does not capture heteroscedasticity.

\section{Designing kernels for interpretability}
\label{sec:design}

The span of the language of kernels used by \procedurename{} is similar to those explored by \citet{DuvLloGroetal13} and \citet{kronberger2013evolution}.
However, \procedurename{} uses a different set of base kernels which are chosen to significantly improve the interpretability of the models produced by our method which we now discuss.

\paragraph{Removal of rational quadratic kernel}

The rational quadratic kernel \citep[e.g.][]{rasmussen38gaussian} can be expressed as a mixture of infinitely many $\kSE$ kernels.
This can have the unattractive property of capturing both long term trends and short term variation in one component.

\begin{figure}[ht]
\centering
\begin{tikzpicture}[]
  \begin{scope}[xshift=0.0\textwidth]
    \node [mybox] (box){
      \fbox{\includegraphics[trim=0cm 0cm 0cm 0cm, clip, width=0.35\columnwidth]{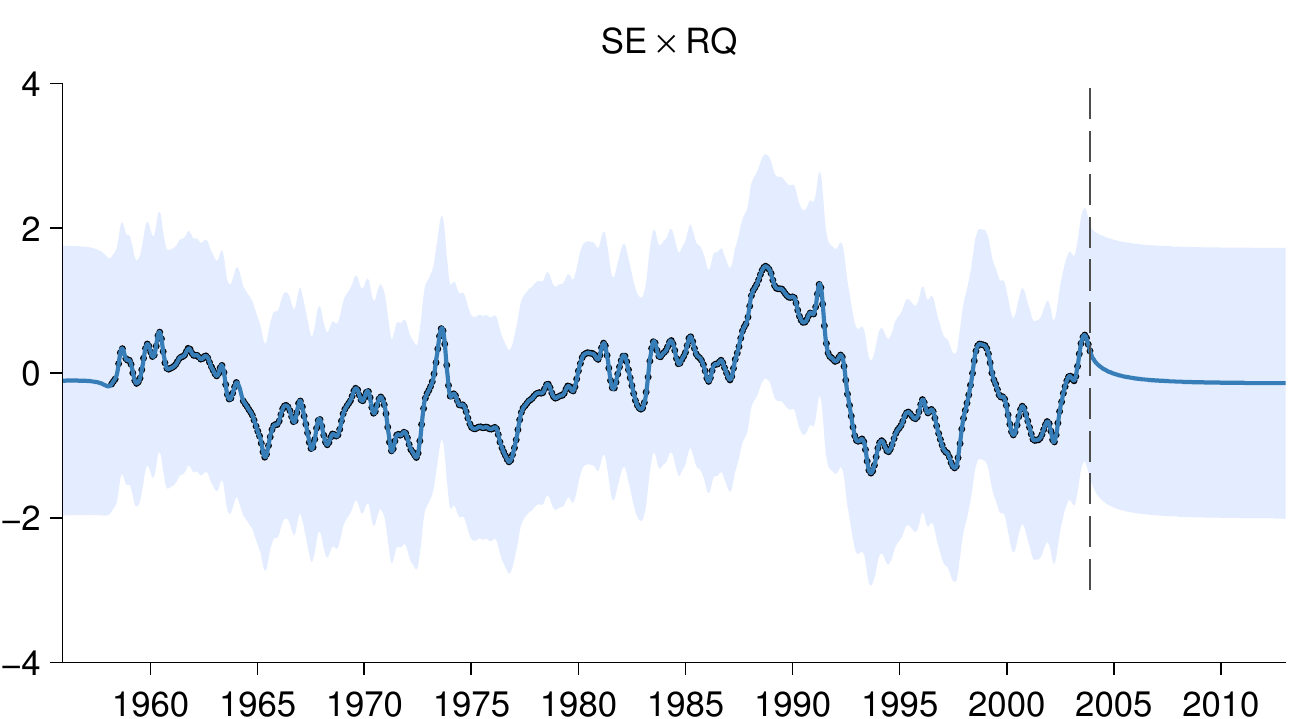}}
    };
  \end{scope}  
  \begin{scope}[xshift=0.25\columnwidth]
  \draw[->,thick] (-0.04\columnwidth,0.05\columnwidth) -- (0.04\columnwidth,0.11\columnwidth); 
  \draw[->,thick] (-0.04\columnwidth,-0.05\columnwidth) -- (0.04\columnwidth,-0.11\columnwidth); 
  \end{scope}
  \begin{scope}[xshift=0.5\columnwidth]
    \begin{scope}[yshift=+0.15\columnwidth]
    \node [mybox] (box){
      \fbox{\includegraphics[trim=0cm 0cm 0cm 0cm, clip, width=0.35\columnwidth]{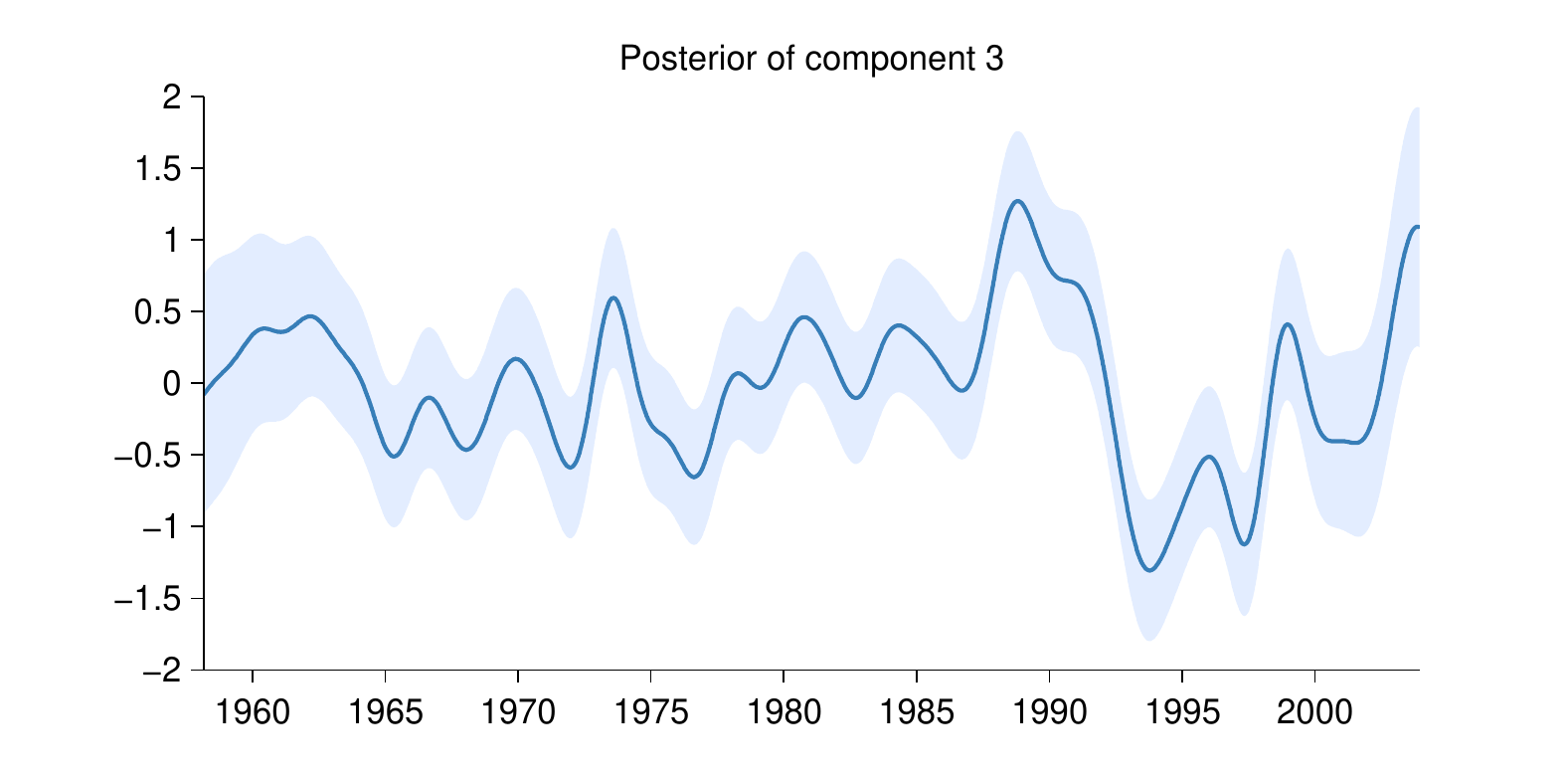}}
    };
    \end{scope}
    \node[font=\LARGE] at (0,0) {$+$};
    \begin{scope}[yshift=-0.15\columnwidth]
    \node [mybox] (box){
      \fbox{\includegraphics[trim=0cm 0cm 0cm 0cm, clip, width=0.35\columnwidth]{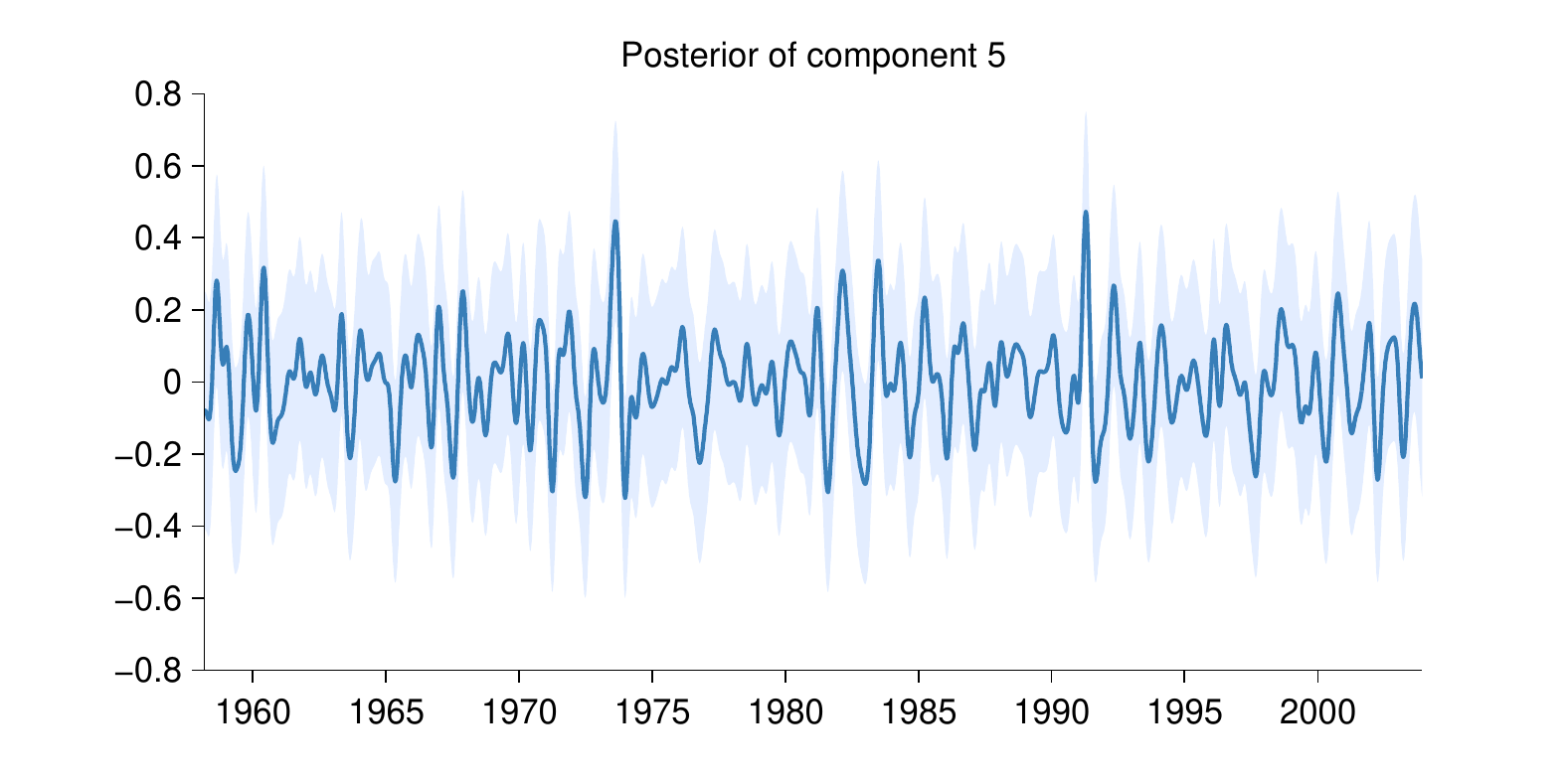}}
    };
    \end{scope}
  \end{scope}
\end{tikzpicture}
\caption{
Left: Posterior of rational quadratic component of model for Mauna Loa data from \citet{DuvLloGroetal13}.
Right: Posterior of two components found by \procedurename{} - the different lenthscales have been separated.
}
\label{fig:rq}
\end{figure}
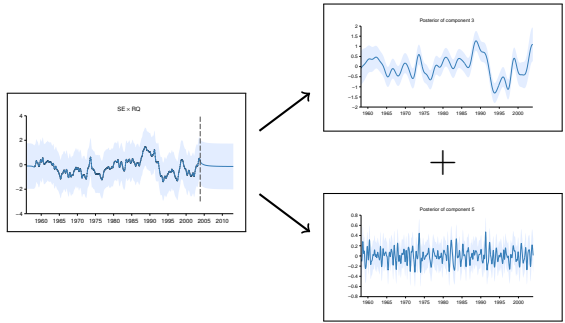

The left of figure~\ref{fig:rq} shows the posterior of a component involving a rational quadratic kernel produced by the procedure of \citet{DuvLloGroetal13} on the Mauna Loa data set (see supplementary material).
This component has captured both a medium term trend and short term variation.
This is both visually unappealing and difficult to describe simply.
In contrast, the right of figure~\ref{fig:rq} shows two of the components produced by \procedurename{} on the same data set which clearly separate the medium term trend and short term deviations.

We do not include the Mat\'ern kernel \citep[e.g.][]{rasmussen38gaussian} used by \citet{kronberger2013evolution} for similar reasons.

\paragraph{Subtraction of unnecessary constants}

The typical definition of the periodic kernel \citep[e.g.][]{rasmussen38gaussian} used by \citet{DuvLloGroetal13} and \citet{kronberger2013evolution} is always greater than zero.
This is not necessary for the kernel to be positive semidefinite; we can subtract a constant from this kernel.
Similarly, the linear kernel used by \citet{DuvLloGroetal13} contained a constant term that can be subtracted.

\begin{figure}[ht]
\centering
\begin{tikzpicture}[]
  \begin{scope}[xshift=0.0\textwidth]
    \begin{scope}[yshift=+0.15\columnwidth]
    \node [mybox] (box){
      \fbox{\includegraphics[trim=0cm 0cm 0cm 0cm, clip, width=0.35\columnwidth]{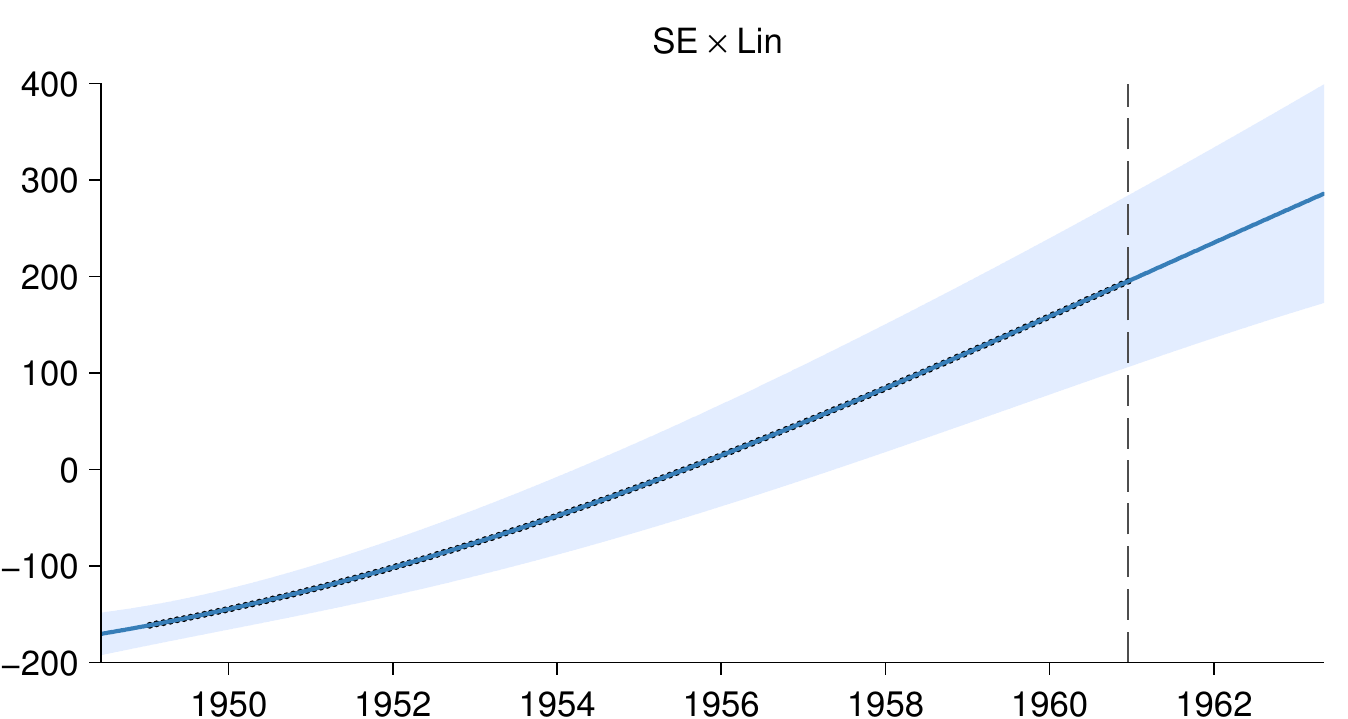}}
    };
    \end{scope}
    \node[font=\LARGE] at (0,0) {$+$};
    \begin{scope}[yshift=-0.15\columnwidth]
    \node [mybox] (box){
      \fbox{\includegraphics[trim=0cm 0cm 0cm 0cm, clip, width=0.35\columnwidth]{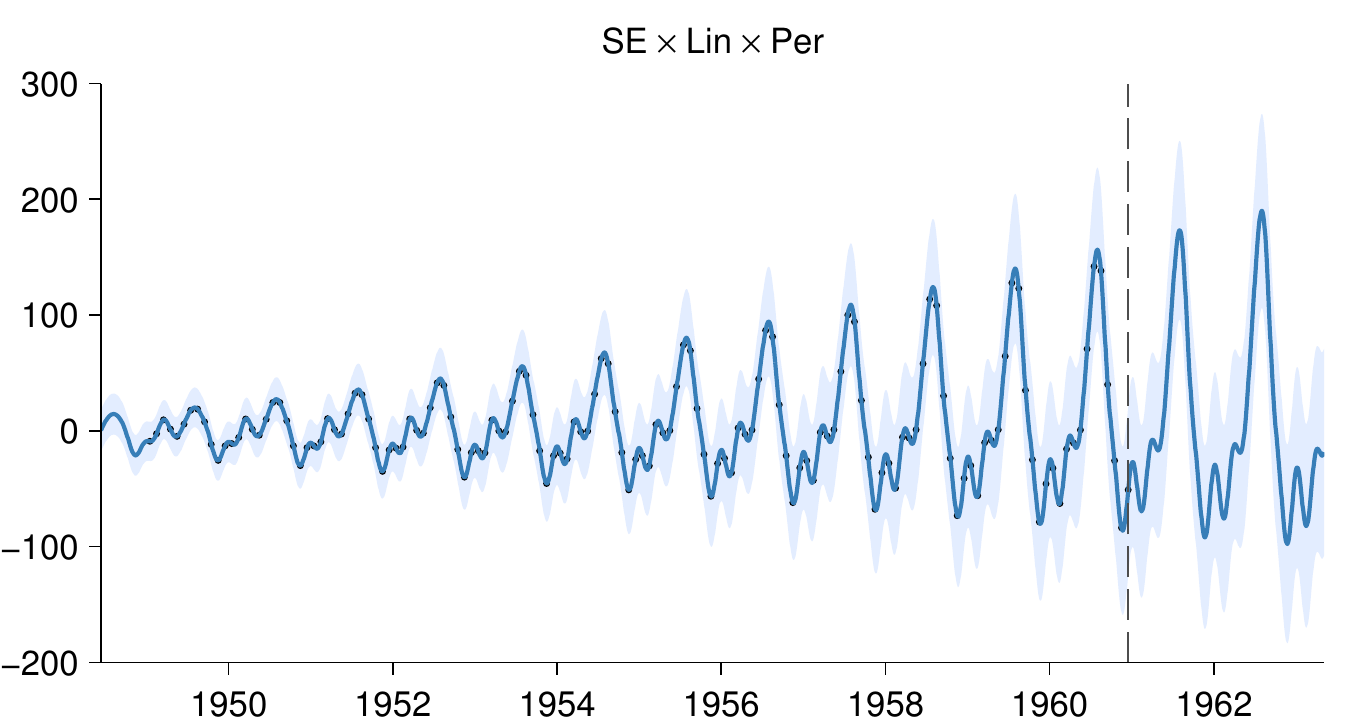}}
    };
    \end{scope}
  \end{scope}  
  \begin{scope}[xshift=0.25\columnwidth]
  \draw[->,thick] (-0.04\columnwidth,0.15\columnwidth) -- (0.04\columnwidth,0.15\columnwidth); 
  \draw[->,thick] (-0.04\columnwidth,-0.15\columnwidth) -- (0.04\columnwidth,-0.15\columnwidth); 
  \end{scope}
  \begin{scope}[xshift=0.5\columnwidth]
    \begin{scope}[yshift=+0.15\columnwidth]
    \node [mybox] (box){
      \fbox{\includegraphics[trim=0cm 0cm 0cm 0cm, clip, width=0.35\columnwidth]{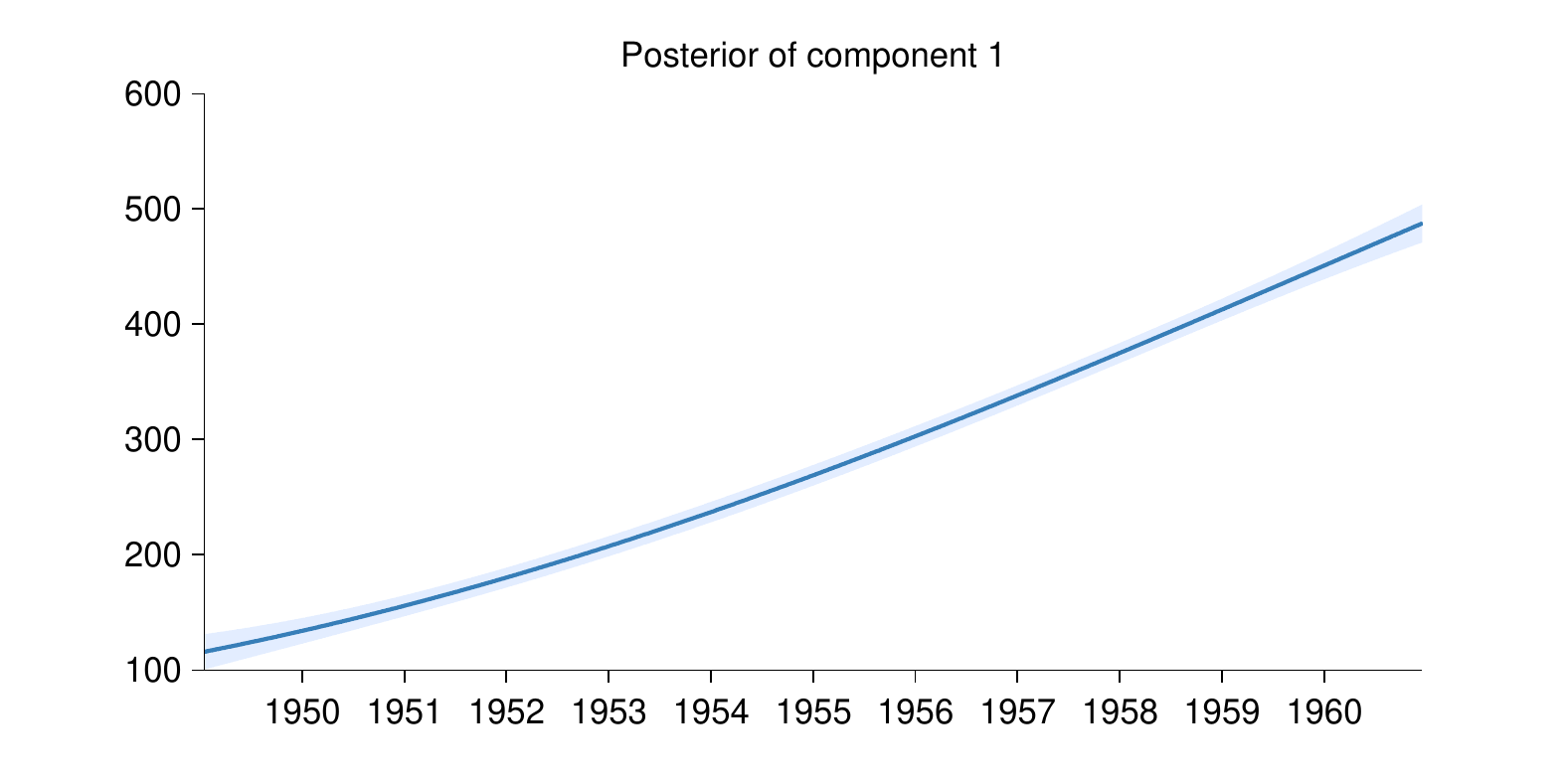}}
    };
    \end{scope}
    \node[font=\LARGE] at (0,0) {$+$};
    \begin{scope}[yshift=-0.15\columnwidth]
    \node [mybox] (box){
      \fbox{\includegraphics[trim=0cm 0cm 0cm 0cm, clip, width=0.35\columnwidth]{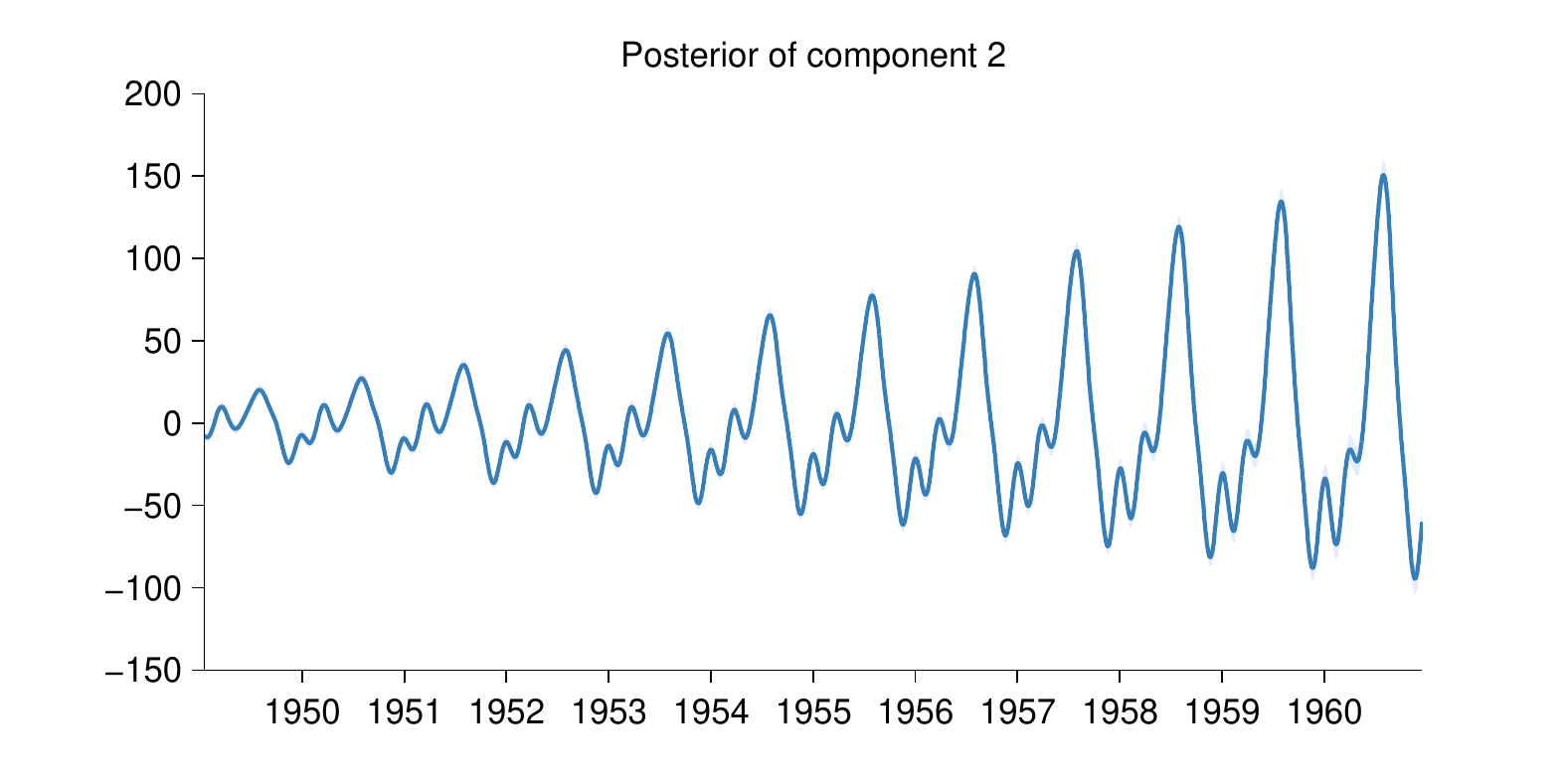}}
    };
    \end{scope}
  \end{scope}
\end{tikzpicture}
\caption{
Left: Posterior of first two components for the airline passenger data from \citet{DuvLloGroetal13}.
Right: Posterior of first two components found by \procedurename{} - removing the constants from $\kLin$ and $\kPer$ has removed the inflated credible intervals due to anti-correlation in the posterior.
}
\label{fig:constant}
\end{figure}
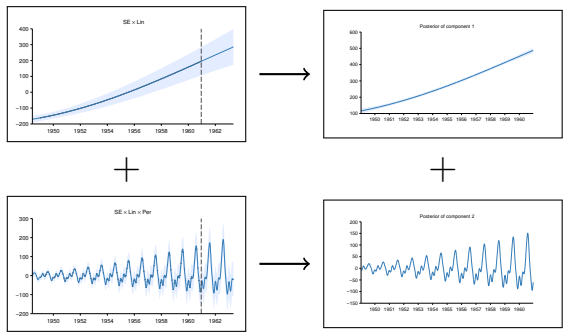

If we had not subtracted these constant, we would have observed two main problems.
First, descriptions of products would become convoluted \eg $(\kPer + \kC) \times (\kLin + \kC) = \kC + \kPer + \kLin + \kPer \times \kLin$ is a sum of four qualitatively different functions.
Second, the constant functions can result in anti-correlation between components in the posterior, resulting in inflated credible intervals for each component which is shown in figure~\ref{fig:constant}.

\section{Related work}
\label{sec:related-work}

\paragraph{Building Kernel Functions}
\cite{rasmussen38gaussian} devote 4 pages to manually constructing a composite kernel to model a time series of carbon dioxode concentrations.
In the supplementary material, we include a report automatically generated by \procedurename{} for this dataset; our procedure chose a model similar to the one they constructed by hand.
Other examples of papers whose main contribution is to manually construct and fit a composite \gp{} kernel are \cite{klenske2012nonparametric} and \cite{lloydgefcom2012}.

\citet{diosan2007evolving, bing2010gp} and \citet{kronberger2013evolution} search over a similar space of models as \procedurename{} using genetic algorithms but do not interpret the resulting models.
Our procedure is based on the model construction method of \citet{DuvLloGroetal13} which automatically decomposed models but components were interpreted manually and the space of models searched over was smaller than that in this work.

\paragraph{Kernel Learning}

Sparse spectrum \gp{}s \citep{lazaro2010sparse} approximate the spectral density of a stationary kernel function using delta functions; this corresponds to kernels of the form $\sum \cos$.
Similarly, \citet{WilAda13} introduce spectral mixture kernels which approximate the spectral density using a scale-location mixture of Gaussian distributions corresponding to kernels of the form $\sum \kSE \times \cos$.
Both demonstrate, using Bochner's theorem \citep{bochner1959lectures}, that these kernels can approximate any stationary covariance function.
Our language of kernels includes both of these kernel classes (see table~\ref{table:motifs}).

There is a large body of work attempting to construct rich kernels through a weighted sum of base kernels called multiple kernel learning (MKL) \citep[e.g.][]{bach2004multiple}.
These approaches find the optimal solution in polynomial time but only if the component kernels and parameters are pre-specified.
We compare to a Bayesian variant of MKL in section~\ref{sec:numerical} which is expressed as a restriction of our language of kernels.

\paragraph{Equation learning}
\cite{todorovski1997declarative}, \cite{washio1999discovering} and \cite{schmidt2009distilling} learn parametric forms of functions specifying time series, or relations between quantities.
In contrast, \procedurename{} learns a parametric form for the covariance, allowing it to model functions without a simple parametric form.

\paragraph{Searching over open-ended model spaces}

This work was inspired by previous successes at searching over open-ended model spaces: matrix decompositions \citep{grosse2012exploiting} and graph structures \citep{kemp2008discovery}.
In both cases, the model spaces were defined compositionally through a handful of components and operators, and models were selected using criteria which trade off model complexity and goodness of fit.
Our work differs in that our procedure automatically interprets the chosen model, making the results accessible to non-experts.

\paragraph{Natural-language output}
To the best of our knowledge, our procedure is the first example of automatic description of nonparametric statistical models.
However, systems with natural language output have been built in the areas of video interpretation \citep{barbu2012video} and automated theorem proving \citep{GanesalingamG13}.

\section{Predictive Accuracy}
\label{sec:numerical}

\begin{figure*}[ht]
\centering
\includegraphics[width=\textwidth]{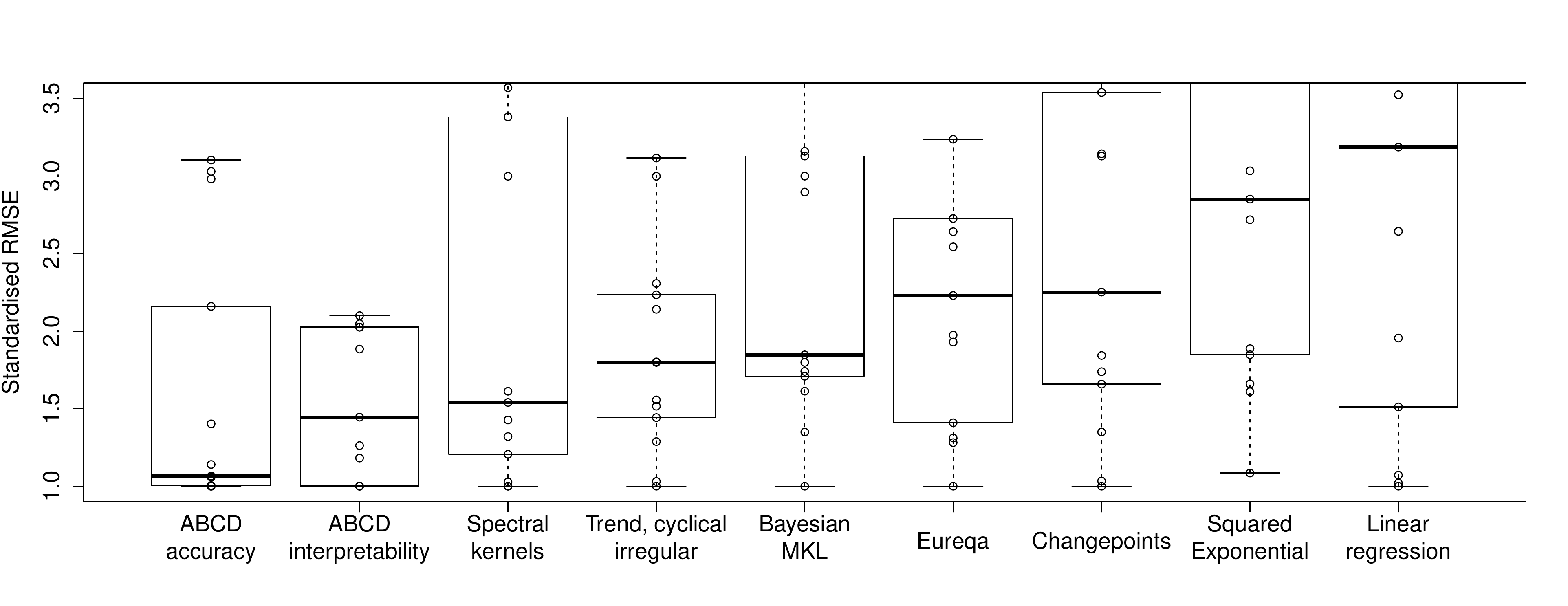}
\vspace{-0.8cm}
\caption{
Raw data, and box plot (showing median and quartiles) of standardised extrapolation RMSE (best performance = 1) on 13 time-series.
The methods are ordered by median.
}
\label{fig:box_extrap_dist}
\end{figure*}

In addition to our demonstration of the interpretability of \procedurename{}, we compared the predictive accuracy of various model-building algorithms at interpolating and extrapolating time-series.
\procedurename{} outperforms the other methods on average.

\paragraph{Data sets}

We evaluate the performance of the algorithms listed below on 13 real time-series from various domains from the time series data library \citep{TSDL}; plots of the data can be found at the beginning of the reports in the supplementary material.

\paragraph{Algorithms}

We compare \procedurename{} to equation learning using Eureqa \citep{Eureqa} and six other regression algorithms: linear regression, \gp{} regression with a single $\kSE$ kernel (squared exponential), a Bayesian variant of multiple kernel learning (MKL) \citep[e.g.][]{bach2004multiple}, change point modeling \citep[e.g.][]{garnett2010sequential, saatcci2010gaussian, FoxDunson:NIPS2012}, spectral mixture kernels \citep{WilAda13} (spectral kernels) and trend-cyclical-irregular models \citep[e.g.][]{lind2006basic}.

\procedurename{} is based on the work of \citet{DuvLloGroetal13}, but with a focus on producing interpretable models.
As noted in section~\ref{sec:design}, the spans of the languages of kernels of these two methods are very similar.
Consequently their predictive accuracy is nearly identical so we only include \procedurename{} in the results for brevity.
Experiments using the genetic programming method of \citet{kronberger2013evolution} are ongoing.

We use the default mean absolute error criterion when using Eureqa.
All other algorithms can be expressed as restrictions of our modeling language (see table~\ref{table:motifs}) so we perform inference using the same search methodology and selection criterion\footnotemark~with appropriate restrictions to the language.
For MKL, trend-cyclical-irregular and spectral kernels, the greedy search procedure of \procedurename{} corresponds to a forward-selection algorithm.
For squared exponential and linear regression the procedure corresponds to marginal likelihood optimisation.
More advanced inference methods are typically used for changepoint modeling but we use the same inference method for all algorithms for comparability.
\footnotetext{We experimented with using unpenalised marginal likelihood as the search criterion but observed overfitting, as is to be expected.} 

We restricted to regression algorithms for comparability; this excludes models which regress on previous values of times series, such as autoregressive or moving-average models \citep[e.g.][]{box2013time}.
Constructing a language for this class of time-series model would be an interesting area for future research.

\paragraph{Interpretability versus accuracy}

BIC trades off model fit and complexity by penalizing the number of parameters in a kernel expression.
This can result in \procedurename{} favoring kernel expressions with nested products of sums, producing descriptions involving many additive components.
While these models have good predictive performance the large number of components can make them less interpretable.
We experimented with distributing all products over addition during the search, causing models with many additive components to be more heavily penalized by BIC.
We call this procedure \procedurename{}-interpretability, in contrast to the unrestricted version of the search, \procedurename{}-accuracy.

\paragraph{Extrapolation}

To test extrapolation we trained all algorithms on the first 90\% of the data, predicted the remaining 10\% and then computed the root mean squared error (RMSE).
The RMSEs are then standardised by dividing by the smallest RMSE for each data set so that the best performance on each data set will have a value of 1.

Figure~\ref{fig:box_extrap_dist} shows the standardised RMSEs across algorithms.
\procedurename{}-accuracy outperforms \procedurename{}-interpretability but both versions have lower quartiles than all other methods.

Overall, the model construction methods with greater capacity perform better: \procedurename{} outperforms trend-cyclical-irregular, which outperforms Bayesian MKL, which outperforms squared exponential.
Despite searching over a rich model class, Eureqa performs relatively poorly, since very few datasets are parsimoniously explained by a parametric equation.

Not shown on the plot are large outliers for spectral kernels, Eureqa, squared exponential and linear regression with values of 11, 493, 22 and 29 respectively.
All of these outliers occurred on a data set with a large discontinuity (see the call centre data in the supplementary material).

\paragraph{Interpolation}
To test the ability of the methods to interpolate, we randomly divided each data set into equal amounts of training data and testing data.
The results are similar to those for extrapolation and are included in the supplementary material.

\section{Conclusion}

Towards the goal of automating statistical modeling we have presented a system which constructs an appropriate model from an open-ended language and automatically generates detailed reports that describe patterns in the data captured by the model.
We have demonstrated that our procedure can discover and describe a variety of patterns on several time series.
Our procedure's extrapolation and interpolation performance on time-series are state-of-the-art compared to existing model construction techniques.
We believe this procedure has the potential to make powerful statistical model-building techniques accessible to non-experts.

\section{Acknowledgements}

We thank Colorado Reed, Yarin Gal and Christian Steinruecken for helpful discussions.
This work was funded in part by NSERC, EPSRC and Google.

\paragraph{Source Code}
Source code to perform all experiments is available on github\footnotemark.
\footnotetext{\url{http://www.github.com/jamesrobertlloyd/gpss-research}. All \gp{} parameter optimisation was performed by automated calls to the GPML toolbox available at \url{http://www.gaussianprocess.org/gpml/code/}.}


\newpage

\begin{appendices}

\section{Kernels}

\subsection{Base kernels}

For scalar-valued inputs, the white noise ($\kWN$), constant ($\kC$), linear ($\kLin$), squared exponential ($\kSE$), and periodic kernels ($\kPer$) are defined as follows:
\begin{eqnarray}
\kWN(\inputVar, \inputVar') =& \sigma^2\delta_{\inputVar, \inputVar'} \\
\kC(\inputVar, \inputVar') =& \sigma^2 \\
\kLin(\inputVar, \inputVar') =& \sigma^2(\inputVar - \ell)(\inputVar' - \ell) \\
\kSE(\inputVar, \inputVar') =& \sigma^2\exp\left(-\frac{(\inputVar - \inputVar')^2}{2\ell^2}\right) \\
\kPer(\inputVar, \inputVar') =&  \sigma^2\frac{\exp\left(\frac{\cos\frac{2 \pi (\inputVar - \inputVar')}{p}}{\ell^2}\right) - I_0\left(\frac{1}{\ell^2}\right)}{\exp\left(\frac{1}{\ell^2}\right) - I_0\left(\frac{1}{\ell^2}\right)}
\end{eqnarray}
where $\delta_{\inputVar, \inputVar'}$ is the Kronecker delta function, $I_0$ is the modified Bessel function of the first kind of order zero and other symbols are parameters of the kernel functions.

\subsection{Changepoints and changewindows}

The changepoint, $\kCP(\cdot,\cdot)$ operator is defined as follows:
\begin{align}
\kCP(\kernel_1, \kernel_2)(x, x') = \qquad \qquad \sigma(x) & k_1(x,x')\sigma(x') \nonumber \\ + (1-\sigma(x)) & k_2(x,x')(1-\sigma(x'))
\end{align}
where $\sigma(x) = 0.5 \times (1 + \tanh(\frac{\ell - x}{s}))$.
This can also be written as
\begin{align}
\kCP(\kernel_1, \kernel_2) = \boldsymbol\sigma\kernel_1 + \boldsymbol{\bar\sigma}\kernel_2
\end{align}
where $\boldsymbol\sigma(x,x') = \sigma(x)\sigma(x')$ and $\boldsymbol{\bar\sigma}(x,x') = (1-\sigma(x))(1-\sigma(x'))$.

Changewindow, $\kCW(\cdot,\cdot)$, operators are defined similarly by replacing the sigmoid, $\sigma(x)$, with a product of two sigmoids.

\subsection{Properties of the periodic kernel}

A simple application of l'H\^opital's rule shows that
\begin{equation}
\kPer(x, x') \to \sigma^2\cos\left(\frac{2 \pi (x - x')}{p}\right) \quad \textrm{as} \quad\ell \to \infty.
\end{equation}
This limiting form is written as the cosine kernel ($\cos$).

\section{Model construction / search}

\subsection{Overview}

The model construction phase of \procedurename{} starts with the kernel equal to the noise kernel, $\kWN$.
New kernel expressions are generated by applying search operators to the current kernel.
When new base kernels are proposed by the search operators, their parameters are randomly initialised with several restarts.
Parameters are then optimized by conjugate gradients to maximise the likelihood of the data conditioned on the kernel parameters.
The kernels are then scored by the Bayesian information criterion and the top scoring kernel is selected as the new kernel.
The search then proceeds by applying the search operators to the new kernel \ie this is a greedy search algorithm.

In all experiments, 10 random restarts were used for parameter initialisation and the search was run to a depth of 10.

\subsection{Search operators}

\procedurename{} is based on a search algorithm which used the following search operators
\begin{eqnarray}
\mathcal{S} &\to& \mathcal{S} + \mathcal{B} \\
\mathcal{S} &\to& \mathcal{S} \times \mathcal{B} \\
\mathcal{B} &\to& \mathcal{B'}
\end{eqnarray}
where $\mathcal{S}$ represents any kernel subexpression and $\mathcal{B}$ is any base kernel within a kernel expression \ie the search operators represent addition, multiplication and replacement.

To accommodate changepoint/window operators we introduce the following additional operators
\begin{eqnarray}
\mathcal{S} &\to& \kCP(\mathcal{S},\mathcal{S}) \\
\mathcal{S} &\to& \kCW(\mathcal{S},\mathcal{S}) \\
\mathcal{S} &\to& \kCW(\mathcal{S},\kC) \\
\mathcal{S} &\to& \kCW(\kC,\mathcal{S})
\end{eqnarray}
where $\kC$ is the constant kernel.
The last two operators result in a kernel only applying outside or within a certain region.

Based on experience with typical paths followed by the search algorithm we introduced the following operators
\begin{eqnarray}
\mathcal{S} &\to& \mathcal{S} \times (\mathcal{B} + \kC)\\
\mathcal{S} &\to& \mathcal{B}\\
\mathcal{S} + \mathcal{S'} &\to& \mathcal{S}\\
\mathcal{S} \times \mathcal{S'} &\to& \mathcal{S}
\end{eqnarray}
where $\mathcal{S'}$ represents any other kernel expression.
Their introduction is currently not rigorously justified.

\section{Predictive accuracy}

\paragraph{Interpolation}

To test the ability of the methods to interpolate, we randomly divided each data set into equal amounts of training data and testing data.
We trained each algorithm on the training half of the data, produced predictions for the remaining half and then computed the root mean squared error (RMSE).
The values of the RMSEs are then standardised by dividing by the smallest RMSE for each data set \ie the best performance on each data set will have a value of 1.

\begin{figure*}[ht]
\centering
\includegraphics[width=\textwidth]{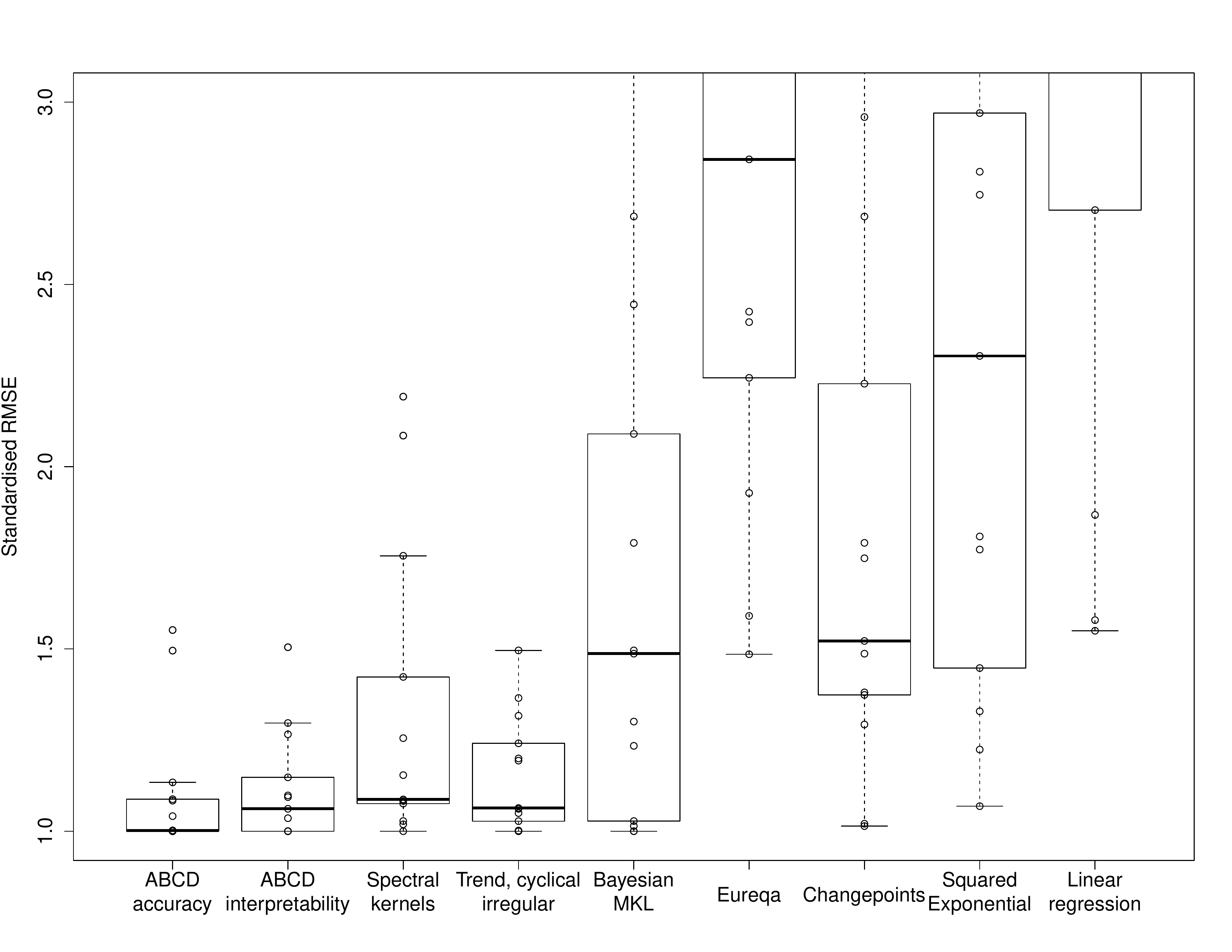}
\caption{
Box plot of standardised RMSE (best performance = 1) on 13 interpolation tasks.
}
\label{fig:box_interp}
\end{figure*}

Figure~\ref{fig:box_interp} shows the standardised RMSEs for the different algorithms.
The box plots show that all quartiles of the distribution of standardised RMSEs are lower for both versions of \procedurename{}.
The median for \procedurename{}-accuracy is 1; it is the best performing algorithm on 7 datasets.
The largest outliers of \procedurename{} and spectral kernels are similar in value.

Changepoints performs slightly worse than MKL despite being strictly more general than Changepoints.
The introduction of changepoints allows for more structured models, but it introduces parametric forms into the regression models (\ie the sigmoids expressing the changepoints).
This results in worse interpolations at the locations of the change points, suggesting that a more robust modeling language would require a more flexible class of changepoint shapes or improved inference (\eg fully Bayesian inference over the location and shape of the changepoint).

Eureqa is not suited to this task and performs poorly.
The models learned by Eureqa tend to capture only broad trends of the data since the fine details are not well explained by parametric forms.

\subsection{Tabels of standardised RMSEs}

See table~\ref{table:interp} for raw interpolation results and table~\ref{table:extrap} for raw extrapolation results. 
The rows follow the order of the datasets in the rest of the supplementary material.
The following abbreviations are used: \procedurename{}-accuracy (\procedurename{}-acc), \procedurename{}-interpretability ((\procedurename{}-int), Spectral kernels (SP), Trend-cyclical-irregular (TCI), Bayesian MKL (MKL), Eureqa (EL), Changepoints (CP), Squared exponential (SE) and Linear regression (Lin).

\begin{table*}[ht]
\center
\begin{tabular}{|c|c|c|c|c|c|c|c|c|}
\hline
\procedurename{}-acc & \procedurename{}-int & SP & TCI & MKL & EL & CP & SE & Lin \\
\hline
1.04 & 1.00 & 2.09 & 1.32 & 3.20 & 5.30 & 3.25 & 4.87 & 5.01\\
1.00 & 1.27 & 1.09 & 1.50 & 1.50 & 3.22 & 1.75 & 2.75 & 3.26\\
1.00 & 1.00 & 1.09 & 1.00 & 2.69 & 26.20 & 2.69 & 7.93 & 10.74\\
1.09 & 1.04 & 1.00 & 1.00 & 1.00 & 1.59 & 1.37 & 1.33 & 1.55\\
1.00 & 1.06 & 1.08 & 1.06 & 1.01 & 1.49 & 1.01 & 1.07 & 1.58\\
1.50 & 1.00 & 2.19 & 1.37 & 2.09 & 7.88 & 2.23 & 6.19 & 7.36\\
1.55 & 1.50 & 1.02 & 1.00 & 1.00 & 2.40 & 1.52 & 1.22 & 6.28\\
1.00 & 1.30 & 1.26 & 1.24 & 1.49 & 2.43 & 1.49 & 2.30 & 3.20\\
1.00 & 1.09 & 1.08 & 1.06 & 1.30 & 2.84 & 1.29 & 2.81 & 3.79\\
1.08 & 1.00 & 1.15 & 1.19 & 1.23 & 42.56 & 1.38 & 1.45 & 2.70\\
1.13 & 1.00 & 1.42 & 1.05 & 2.44 & 3.29 & 2.96 & 2.97 & 3.40\\
1.00 & 1.15 & 1.76 & 1.20 & 1.79 & 1.93 & 1.79 & 1.81 & 1.87\\
1.00 & 1.10 & 1.03 & 1.03 & 1.03 & 2.24 & 1.02 & 1.77 & 9.97\\
\hline
\end{tabular}
\caption{Interpolation standardised RMSEs}
\label{table:interp}
\end{table*}

\begin{table*}[ht]
\center
\begin{tabular}{|c|c|c|c|c|c|c|c|c|}
\hline
\procedurename{}-acc & \procedurename{}-int & SP & TCI & MKL & EL & CP & SE & Lin \\
\hline
1.14 & 2.10 & 1.00 & 1.44 & 4.73 & 3.24 & 4.80 & 32.21 & 4.94\\
1.00 & 1.26 & 1.21 & 1.03 & 1.00 & 2.64 & 1.03 & 1.61 & 1.07\\
1.40 & 1.00 & 1.32 & 1.29 & 1.74 & 2.54 & 1.74 & 1.85 & 3.19\\
1.07 & 1.18 & 3.00 & 3.00 & 3.00 & 1.31 & 1.00 & 3.03 & 1.02\\
1.00 & 1.00 & 1.03 & 1.00 & 1.35 & 1.28 & 1.35 & 2.72 & 1.51\\
1.00 & 2.03 & 3.38 & 2.14 & 4.09 & 6.26 & 4.17 & 4.13 & 4.93\\
2.98 & 1.00 & 11.04 & 1.80 & 1.80 & 493.30 & 3.54 & 22.63 & 28.76\\
3.10 & 1.88 & 1.00 & 2.31 & 3.13 & 1.41 & 3.13 & 8.46 & 4.31\\
1.00 & 2.05 & 1.61 & 1.52 & 2.90 & 2.73 & 3.14 & 2.85 & 2.64\\
1.00 & 1.45 & 1.43 & 1.80 & 1.61 & 1.97 & 2.25 & 1.08 & 3.52\\
2.16 & 2.03 & 3.57 & 2.23 & 1.71 & 2.23 & 1.66 & 1.89 & 1.00\\
1.06 & 1.00 & 1.54 & 1.56 & 1.85 & 1.93 & 1.84 & 1.66 & 1.96\\
3.03 & 4.00 & 3.63 & 3.12 & 3.16 & 1.00 & 5.83 & 5.35 & 4.25\\
\hline
\end{tabular}
\caption{Extrapolation standardised RMSEs}
\label{table:extrap}
\end{table*}

\section{Guide to the automatically generated reports}

Additional supplementary material to this paper is 13 reports automatically generated by \procedurename{}.
A link to these reports will be maintained at \url{http://mlg.eng.cam.ac.uk/lloyd/}.
We recommend that you read the report for `01-airline' first and review the reports that follow afterwards more briefly.
`02-solar' is discussed in the main text.
`03-mauna' analyses a dataset mentioned in the related work.
`04-wheat' demonstrates changepoints being used to capture heteroscedasticity.
`05-temperature' extracts an exactly periodic pattern from noisy data.
`07-call-centre' demonstrates a large discontinuity being modeled by a changepoint.
`10-sulphuric' combines many changepoints to create a highly structured model of the data.
`12-births' discovers multiple periodic components.

\end{appendices}

\newpage

\bibliography{gpss}
\bibliographystyle{format/aaai}

\end{document}